\documentclass[10pt,twocolumn]{article}

\usepackage{graphicx}
\usepackage{microtype}
\usepackage{subfigure}
\usepackage{booktabs} 

\usepackage{hyperref}

\graphicspath{ {fig/} }

\usepackage{amsmath}
\usepackage{amssymb}
\usepackage{mathtools, cuted}	
\usepackage[toc,]{appendix}

\usepackage{tikz}
\usetikzlibrary{er,positioning,shapes,snakes,shadows,arrows, topaths,matrix,arrows,fit,intersections,shapes.arrows}

\makeatletter
\pgfkeys{/pgf/.cd,
  parallelepiped offset x/.initial=2mm,
  parallelepiped offset y/.initial=2mm
}
\pgfdeclareshape{parallelepiped}
{
  \inheritsavedanchors[from=rectangle] 
  \inheritanchorborder[from=rectangle]
  \inheritanchor[from=rectangle]{north}
  \inheritanchor[from=rectangle]{north west}
  \inheritanchor[from=rectangle]{north east}
  \inheritanchor[from=rectangle]{center}
  \inheritanchor[from=rectangle]{west}
  \inheritanchor[from=rectangle]{east}
  \inheritanchor[from=rectangle]{mid}
  \inheritanchor[from=rectangle]{mid west}
  \inheritanchor[from=rectangle]{mid east}
  \inheritanchor[from=rectangle]{base}
  \inheritanchor[from=rectangle]{base west}
  \inheritanchor[from=rectangle]{base east}
  \inheritanchor[from=rectangle]{south}
  \inheritanchor[from=rectangle]{south west}
  \inheritanchor[from=rectangle]{south east}
  \backgroundpath{
    \southwest \pgf@xa=\pgf@x \pgf@ya=\pgf@y
    \northeast \pgf@xb=\pgf@x \pgf@yb=\pgf@y
    \pgfmathsetlength\pgfutil@tempdima{\pgfkeysvalueof{/pgf/parallelepiped offset x}}
    \pgfmathsetlength\pgfutil@tempdimb{\pgfkeysvalueof{/pgf/parallelepiped offset y}}
    \def\ppd@offset{\pgfpoint{\pgfutil@tempdima}{\pgfutil@tempdimb}}
    \pgfpathmoveto{\pgfqpoint{\pgf@xa}{\pgf@ya}}
    \pgfpathlineto{\pgfqpoint{\pgf@xb}{\pgf@ya}}
    \pgfpathlineto{\pgfqpoint{\pgf@xb}{\pgf@yb}}
    \pgfpathlineto{\pgfqpoint{\pgf@xa}{\pgf@yb}}
    \pgfpathclose
    \pgfpathmoveto{\pgfqpoint{\pgf@xb}{\pgf@ya}}
    \pgfpathlineto{\pgfpointadd{\pgfpoint{\pgf@xb}{\pgf@ya}}{\ppd@offset}}
    \pgfpathlineto{\pgfpointadd{\pgfpoint{\pgf@xb}{\pgf@yb}}{\ppd@offset}}
    \pgfpathlineto{\pgfpointadd{\pgfpoint{\pgf@xa}{\pgf@yb}}{\ppd@offset}}
    \pgfpathlineto{\pgfqpoint{\pgf@xa}{\pgf@yb}}
    \pgfpathmoveto{\pgfqpoint{\pgf@xb}{\pgf@yb}}
    \pgfpathlineto{\pgfpointadd{\pgfpoint{\pgf@xb}{\pgf@yb}}{\ppd@offset}}
  }
}
\makeatother

\usepackage{xcolor}
\numberwithin{equation}{section}

\usepackage{natbib}

\usepackage{algorithm}
\usepackage{algorithmic}

\begin{document}
\title{Generating new pictures in complex datasets with a simple neural network}
\author{Galin Georgiev\\ GammaDynamics, LLC\footnote{galin.georgiev@gammadynamics.com}}
\date{}
\maketitle

\vskip 0.3in

\begin{abstract}
We introduce a version of a variational auto-encoder (VAE), which can generate good perturbations of images, when trained on a complex dataset (in our experiments, CIFAR-10). The net is using only two latent generative dimensions per class, with uni-modal probability density. The price one has to pay for good generation is that not all training images are well reconstructed. An additional classifier is required to determine which training image is well reconstructed and generally the weights of training images.  Only training images which are well reconstructed, can be perturbed. For good perturbations, we use the tentative empirical drifts of well reconstructed images. The construct is not predictive in the usual statistical sense.
\end{abstract}

\section{Introduction}
\label{introduction}
Generative networks not only re-create observations, i.e., create whole probability distributions per observation (as opposed to, say, one number), they also promise to create \emph{new}, i.e., not seen before, observations. Currently, there are four major types of generative networks: flow-based\footnote{The term \emph{flow} appears to be used for the first time in this context in \cite{Tabak10}, section 1.}, \cite{Bengio14}, \cite{Kingma18}, auto-regressive \cite{Oord16}, variational auto-encoders (VAE-s), \cite{Rezende14}, \cite{Kingma14} and generative adversarial nets (GAN-s), \cite{Goodfellow14}, \cite{Goodfellow16}. For a full recent taxonomy, see for example \cite{Kingma18}. Because biological brains use only a few dimensions for creativity, we will focus here on generative nets which allow for small dimensions and efficiency of creativity. Of the above types, only VAE-s and GAN-s are parallelizable (i.e., efficient) and allow for a small number of \emph{creative} dimensions (GAN-s allow for a small dimensional \emph{encoder}, see \cite{Ulyanov17}). In our experiments, we used only two dimensions per class, with uni-modal probability density for generation.

Firstly, we propose in sub-section \ref{training observation} to use \emph{weights} of observations, when training the net. As a result, not all training observations are well reconstructed (reconstruction is a necessary condition for perturbation of that image). An additional classifier is required to determine the weights of training observations and hence which training observation is well reconstructed. Because of this choice of training observations, the construct is not predictive in the usual statistical sense.

Secondly, we propose in sub-section \ref{observation-specific measure} to use the tentative empirical drift $\neq 0$ of well reconstructed images for generation, as opposed to drift $=0$, as normally done. The knowledge of the tentative empirical drifts require that the net be optimized at least twice. This makes generative nets \emph{perturbative}, in the sense that actual observation is required. That is how human creativity normally works: to be considered not pathological, it starts with some actual observation and deviates thereafter. 

\section{Method}
\label{method}

We will focus in what remains on variational auto-encoders (VAE-s) and their modifications. The method proposed here can in theory be used with other generative nets with small-dimensional encoder (the other current candidate for small-dimensional parallelizable generative samples -- the generative adversarial nets (GAN-s) with encoder, \cite{Ulyanov17} -- has some theoretical problems when the training set is small, \cite{Arora17-2}).

As all generative nets, VAE-s work in two regimes:
\begin{itemize}
\item \emph{non-creative} regime, with the training observations $\{\mathbf{x}_{\mu}\}$   fed to the input layer of the net. VAE-s random sample in this regime from a closed-form conditional \emph{posterior} model probability density $p(\mathbf{z} | \mathbf{x}_{\mu})$.
\item \emph{creative} regime, with no data clamped onto the net. Random sampling is normally from known probability density $p(\mathbf{z})$ with zero drift. In our case, we random sample using same density but with \emph{tentative empirical non-zero drifts} of training observations, making VAE-s \emph{perturbative} in the sense of \cite{Georgiev15-2}, Section 3. Knowledge of the tentative empirical drifts of the training observations requires to run VAE-s in non-creative regime at least twice!

\end{itemize}
In order to do reconstruction, variational auto-encoders also introduce a conditional model \emph{reconstruction} density $p^{rec}(\mathbf{x}_{\mu} | \mathbf{z})$, for training observations $\{\mathbf{x}_{\mu}\}$.  In non-creative regime, the reconstruction error at the output layer of the net is the expectation $\mathbf{E}(- \log p^{rec}(\mathbf{x}_{\mu}|\mathbf{z}))_{p(\mathbf{z}|\mathbf{x}_{\mu})}$, where $\mathbf{E}_{\phi()}$ stands for \emph{expectation} with respect to measure $\phi()$, with Monte Carlo usually used to compute it in practice. In the creative regime, we have a joint model density $p(\mathbf{x}_{\mu},\mathbf{z})$ $:=p^{rec}(\mathbf{x}_{\mu}|\mathbf{z})p(\mathbf{z})$. The unknown data density  $q(\mathbf{x}_{\mu})$ is the implied marginal: 
\begin{align}
q(\mathbf{x}_{\mu}) =\int p(\mathbf{x}_{\mu},\mathbf{z})d\mathbf{z} = \frac{ p(\mathbf{x}_{\mu},\mathbf{z})}{ q(\mathbf{z}|\mathbf{x}_{\mu}) },
\label{method.1}
\end{align}
for observation $\mathbf{x}_{\mu}$, some implied posterior conditional density $q(\mathbf{z}|\mathbf{x}_{\mu})$ which  is generally intractable, $q(\mathbf{z}|\mathbf{x}_{\mu})$ $\neq p(\mathbf{z}|\mathbf{x}_{\mu})$. The full decomposition of our minimization target -- the negative log-likelihood of the unknown data density $-\log q(\mathbf{x}_{\mu})$, also called \emph{cross-entropy} -- is easily derived via the Bayes rules, \cite{Georgiev15-2}, section 3:
\begin{align}
-\log q(\mathbf{x}_{\mu}) =
 \underbrace{\mathbf{E}(- \log p^{rec}(\mathbf{x}_{\mu}| \mathbf{z}))_{ p(\mathbf{z}|\mathbf{x}_{\mu})} }_{reconstruction~error} + \nonumber \\
+ \underbrace{\mathcal{D}( p(\mathbf{z| x}_{\mu}) || p(\mathbf{z}))}_{generative~error} - \underbrace{\mathcal{D}( p(\mathbf{z| x}_{\mu}) || q(\mathbf{z| x}_{\mu}) )}_{variational~error},
\label{method.2}
\end{align}
where  $\mathcal{D}( || )$ is the \emph{Kullback-Leibler divergence} and $\mathbf{x}_{\mu}$ is an observation.

The \emph{reconstruction error} measures the negative likelihood of getting $\mathbf{x}_{\mu}$ back, after the transformations and  randomness inside the net. The \emph{generative error} is the  divergence between the generative densities in the non-creative and creative regimes. VAE-s are conceptual in physics sense, because when the generative densities $p()$ are in the exponential/Gibbs class, the generative error satisfies the generalized second principle of thermodynamics - see \cite{Georgiev15-2}, section 2. 

The \emph{variational error} is an approximation error: it is the price for having a tractable generative density $p(\mathbf{z} | \mathbf{x}_{\mu})$ in the non-creative regime.  When the reconstruction density $p^{rec}(\mathbf{x}_{\mu}|\mathbf{z})$ is fixed (near the end of the optimization, for example), the minimization of the generative error implies minimization of the variational error: by definition, $q(\mathbf{z}|\mathbf{x}_{\mu}) = \frac{p^{rec}(\mathbf{x}_{\mu}|\mathbf{z})}{q(\mathbf{x}_{\mu})} p(\mathbf{z})$ and the unknown data density $q(\mathbf{x}_{\mu})$ is fixed. This variational error is ignored in VAE-s and only the respective  upper bound of the cross-entropy (the sum of the remaining two errors) is minimized:
\begin{align}
- \log \mathcal{L}_{VAE}(\mathbf{x}_{\mu}) &=
 \underbrace{\mathbf{E}(- \log p^{rec}(\mathbf{x}_{\mu}| \mathbf{z}))_{ p(\mathbf{z}|\mathbf{x}_{\mu})} }_{reconstruction~error} + \nonumber \\
&+ \underbrace{\mathcal{D}( p(\mathbf{z| x}_{\mu}) || p(\mathbf{z}))}_{generative~error},
\label{method.3}
\end{align}
where $\mathbf{x}_{\mu}$ is a training observation and  $p(\mathbf{z} | \mathbf{x}_{\mu})$, $p(\mathbf{z})$ are, yet to be chosen, closed-form model probability densities. 

When the generative densities $p()$ are in the exponential/Gibbs class and one imposes the VAE minimization goal (\ref{method.3}), the variational error can be computed without the unknown true data posterior conditional density $q(\mathbf{z}|\mathbf{x}_{\mu})$, \cite{Georgiev15-2}, sub-section 3.8.

\subsection{Training observation weight}
\label{training observation}

As far as we know, the first successful attempt to put weights in observations of the likelihood, preserving its nice asymptotic properties, was made in \cite{Hu94} and papers following it  (the weights sum up to the size of the training set). By definition, all likelihood-based generative models can benefit from this generalization of the definition: likelihood is the empirical expectation -- essentially the weighted sum -- of model densities of observations (negative log-likelihood is also known as \emph{cross-entropy}).

In the notations (\ref{method.3}), instead of using as minimization target: $- \sum_{\mu} \log \mathcal{L}_{VAE}(\mathbf{x}_{\mu})$, we propose to use: 

\begin{align}
- \sum_{\mu}\mathcal{W}(\mathbf{x}_{\mu}) \log \mathcal{L} _{VAE}(\mathbf{x}_{\mu}) + \sum_{\mu}\mathcal{W}(\mathbf{x}_{\mu}), 
\label{method.4}
\end{align}
where $\mathbf{x}_{\mu}$ is training observation and the sum of all weights $\mathcal{W}$ over the training set is the size of the training set. 

Because the usual VAE minimization summand (\ref{method.3}) consists of two errors only, and the generative error of VAE-s is usually relatively large, i.e. generation is usually bad, \cite{Rosca18}, something has to give, and this ``something'' is the other error -- the reconstruction error. As a result, not all training observations are well  reconstructed -- see Figure \ref{Fig.Results.1}, \textbf{Middle}. That is how normal biological brain works -- once the number of new observations exceed certain small number, the normal biological brain can not reconstruct new observations very well! The smaller the number of new observations, the easier (see Figures \ref{Fig.Results.0},  \ref{Fig.Results.1}, \ref{Fig.Results.2} and \ref{Fig.Results.3}).

This construct memorizes the training data and is not \emph{predictive} in the usual statistical sense -- the observations not seen in training can not be well reconstructed (see Figure \ref{Fig.Results.1a}). Nevertheless, it generates new images, which are usually variations of well reconstructed training images. It also learns the low-dimensional manifold, where the well reconstructed training images ``live'' (in our experiments, two dimensions per class, with uni-modal probability density)  -  for more on \emph{manifold learning}, see for example \cite{Cayton05}, \cite{Rifai12}. Also, the difference of log-likelihoods between training and testing set (the so-called \emph{generalization error}, \cite{Zhang16}) is normal. As pointed out in \cite{Theis15}, there is no direct relation between likelihoods and quality of generated images.

When separate weight classifier is used, with set maximum ratio between weights, only one observation weight per minibatch (the maximum weight) is typically large and ``survives''. In our experiments for example, the maximum ratio between weights was set to $10^6$. The introduction of additional classifier makes the whole architecture \emph{auto-classifier-encoder} or ACE for short, the variation of which was introduced in \cite{Georgiev15-2}. 

One can of course hard-code the training observations which one wants well reconstructed (which we did in our experiments -- see the Appendix for more details). The rest of the training observations which are well reconstructed (one per minibatch) are chosen by the weight classifier. 

The bottom line is that minibatch size is \emph{inversely proportional} to the number of training observations which are well reconstructed. The larger the minibatch size, the smaller the number of well reconstructed training observations. For example, for 50,000 training observations and minibatch size of 1,000, only $50 = 50,000/1,000$ training observations will be well reconstructed.

\subsection{Observation-specific measure (drift)}
\label{observation-specific measure}

When multiple epochs are run -- as is generally required for proper optimization -- the result is a full Mote Carlo simulation on the used density. If only one random sample per epoch, per observation is used (as is normally the case in VAE-s -- see for example \cite{Kingma14}, section 2), the total number of random samples per observation is then equal to the number of epochs.

Matching the $log$ of the standard deviation $log$ $\sigma,$ although an important part of VAE-s, turns out not as important as matching the drift $\mu,$ when training and sampling. Every training observation has its own drift in training, typically very different from $0$. Although VAE-s reconstruct well training observations, they generate poor new observations (when drift $0$ is used to generate new random observations, as is usually the case, Figure \ref{Fig.Results.4}) - see for example \cite{Hoffman16}, \cite{Dimoulin16}, \cite{Rosca18}.

In the original paper \cite{Black73}, when pricing derivatives of financial instruments, the change of measure/drift to ``risk-neutral'' was used. It is now well-known that Girsanov Theorem, \cite{Girsanov60}, is behind the fact that one can change the drift of Brownian motion to price financial derivatives.

VAE-s are not much different: use of specific training observation tentative drifts $\neq 0$ on the right-hand side of the generative error in (\ref{method.3}) to generate new observations, makes VAE-s \emph{perturbative}, in the sense of needing an observation to wiggle around (\cite{Georgiev15-2}, Section 3). Then the generative error becomes theoretically zero. Biological brains are perturbative: even human painting geniuses use perturbations of exact images to retain some form of recognition.

Using for generation the observation-specific drifts, learned in training, is at first sight counter-intuitive: after all, generation with the same drifts (and different random samples) can theoretically only reproduce the original observations (as long as the same density is used for training and generation)? The way around this is: i) use big number of standard deviations, not encountered in training, to deviate from the original observation (Figure \ref{Fig.Results.0}), especially when the standard deviations learned in training are really small, or ii) not random sample in training at all, i.e. violate the basic tenet in VAE-s.

All VAE-s interpolating between images have to match original images and therefore use that observation-specific drift (see Figure \ref{Fig.Results.5}).

At least one optimization run of VAE is needed to know the tentative empirical drifts of the training observations (see Figures \ref{Fig.Results.0}, \ref{Fig.Results.2} and \ref{Fig.Results.3}). When the tentative empirical (observation-specific) drifts from the first optimization run of VAE-s are used on the right-hand side of the generative error in (\ref{method.3}), the generative error in (\ref{method.3}) becomes theoretically zero (in practice, when the second or greater optimization runs take place, the left-hand side of the generative error changes, so the generative error is slightly above zero, even in convergence).

\section{Results}

We use CIFAR-10 dataset which has 50,000 training and 10,000 testing 32 x 32 pixel colored images, with dimension $3 * 32 * 32$, \cite{Krizhevsky09}.

\begin{figure*}[!ht]
\center
\includegraphics[width=\textwidth]{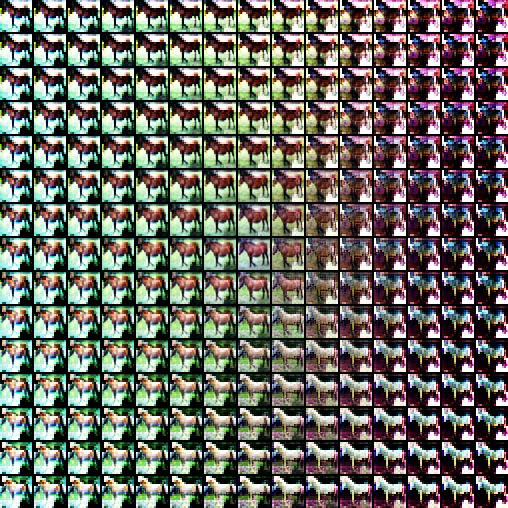}
\caption{New images centered around the 8-th training image in the initial CIFAR-10 order. This image is well reconstructed in the first minibatch i.e. its weight in the likelihood of the resp. first minibatch is hard-coded to be high, as can be seen in Figure \ref{Fig.Results.1}. Using the tentative empirical drift from the second run of our construct ((1.7199,-18.9265) in the first Laplacian pyramid level and (-0.6789,-2.3658) in the second), as explained in sub-section \ref{observation-specific measure}, and standard deviation of zero, the image is decoded and placed at the center. The new images are created using for random samples in horizontal and vertical dimensions 15-strong, equally-spaced, deterministic grid $\{\sigma_{x} \}_{1}^{15}$, $\{\sigma_{y} \}_{1}^{15}$, $-7 \leq \sigma_{x,y} \leq 7$, irrespective of the empirical standard deviation. The rest of the details of model architecture are in the Appendix.}
\label{Fig.Results.0}
\end{figure*}

\begin{figure*}[!ht]
\center
\includegraphics[width=\textwidth]{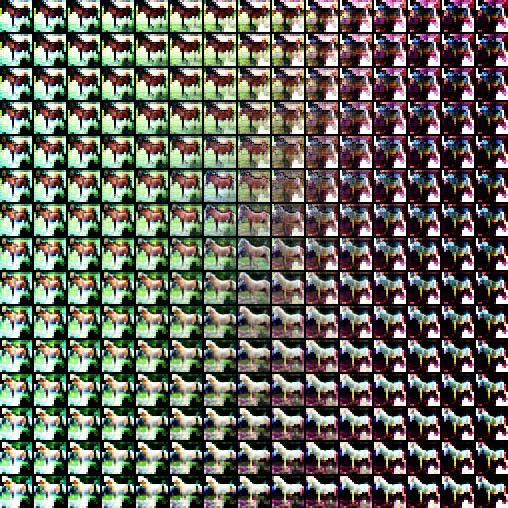}
\caption{New images centered around training image number 1020 in the initial CIFAR-10 order.  Using its tentative empirical drift from the second run of our construct ((-2.8323,-17.2063) in the first Laplacian pyramid level and (2.9199,-0.9004) in the second), as explained in sub-section \ref{observation-specific measure}, and standard deviation of zero, the image is decoded and placed at the center. The rest is as in Figure \ref{Fig.Results.0}.}
\label{Fig.Results.2}
\end{figure*}

\begin{figure*}[!ht]
\center
\includegraphics[width=\textwidth]{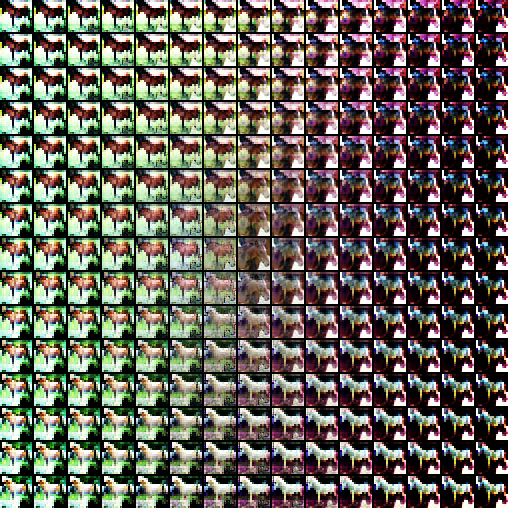}
\caption{New images centered around training image number 2016 in the initial CIFAR-10 order.  Using its tentative empirical drift from the second run of our construct ((1.7418,-13.6416) in the first Laplacian pyramid level and (-0.1847,0.9476) in the second), as explained in sub-section \ref{observation-specific measure}, and standard deviation of zero, the image is decoded and placed at the center. The rest is as in Figure \ref{Fig.Results.0}.}
\label{Fig.Results.3}
\end{figure*}

\begin{figure*}[!ht]
\center
\includegraphics[width=\textwidth]{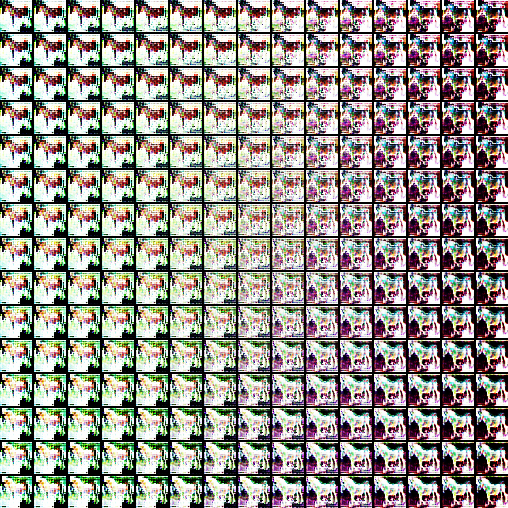}
\caption{New images centered around latent drift of 0, as is usually done when VAE-s are used for generating original images. There is no training image, corresponding to the drift of 0. The rest is as in Figure \ref{Fig.Results.0}. There is not much to be seen, in accordance with the usual complaint that VAE-s have poor generating power when drift of 0 is used - see sub-section \ref{observation-specific measure} and \cite{Rosca18}.}
\label{Fig.Results.4}
\end{figure*}

\begin{figure}[!ht]
\begin{minipage}{\columnwidth}
\includegraphics[width=\columnwidth]{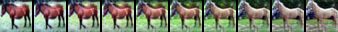}
\end{minipage}
\vskip 0.1in
\begin{minipage}{\columnwidth}
\includegraphics[width=\columnwidth]{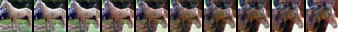}
\end{minipage}  
\vskip 0.1in
\begin{minipage}{\columnwidth}
\includegraphics[width=\columnwidth]{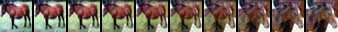}
\end{minipage} \hfill
\caption{ Linear interpolation in the latent space between training images number 8 (leftmost) and 1020 (rightmost) \textbf{Top}, 1020 (leftmost) and 2016 (rightmost) \textbf{Middle}, 8 (leftmost) and 2016 (rightmost) \textbf{Bottom}, in the initial CIFAR-10 order. All images belong to class ``horse'' and only 2 dimensions of latent space, minibatch size of 1,000 and class-specific sampler and decoder are used. In order to match exactly interpolated images, tentative empirical drifts are used from the second run of our construct -- see main text and Figures \ref{Fig.Results.0}, \ref{Fig.Results.2} and \ref{Fig.Results.3}. Only three images were hard-coded to be well reconstructed in the resp. first three minibatches  -- the above 8, 1020, 2016-th training images in the initial CIFAR-10 order; the other $47=50,000/1,000 - 3$ well reconstructed images were chosen by the weight classifier. The rest of the details of model architecture are in the Appendix.}
\label{Fig.Results.5}
\end{figure}

In  Figures \ref{Fig.Results.0}, \ref{Fig.Results.2} and \ref{Fig.Results.3} we present the new images which are variations of well reconstructed training images. In Figure \ref{Fig.Results.5} we present linear interpolations between well reconstructed images in two-dimensional (per class) latent space, with uni-modal probability density.  The model architecture is in the Appendix.

\begin{figure*}[!ht]
\center
\includegraphics[width=\textwidth]{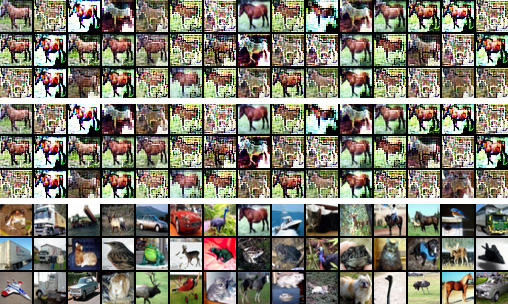}
\caption{The \textbf{Bottom} shows the first 45 raw training images in the initial CIFAR-10 order. The \textbf{Middle} 45 images show \emph{reconstruction} of those raw images. The \textbf{Top} 45 images show \emph{generation}, using different than the resp.  images in the \textbf{Middle} random sample but the same tentative empirical drift (from the second run of our construct). The \textbf{Top} images should be close to the images in the \textbf{Middle}. The minibatch size is 1,000 and only one training image per minibatch gets well-reconstructed i.e. its weight in the  likelihood is hard-coded to be high, as explained in sub-section \ref{training observation} (in this case, the 8-th image in the first training minibatch). As one can see, the 8-th image is well reconstructed in the first training minibatch but not the other images in the minibatch (the counting is from left to right and from top to bottom). The fact that the decoder uses class ``horse'', as opposed to the ``right'' classes, to which training images at the bottom belong, is not an accident. When decoder is class-dependent and there is a batch normalization, \cite{Ioffe15}, there is a dependence on minibatch and class. Then, for hard-coded training images, sub-section \ref{training observation}, one has to use without any loss, for the whole respective minibatch, the class to which hard-coded images belong. The rest of the details of model architecture are in the Appendix.}
\label{Fig.Results.1}
\end{figure*}

\begin{figure*}[!ht]
\center
\includegraphics[width=\textwidth]{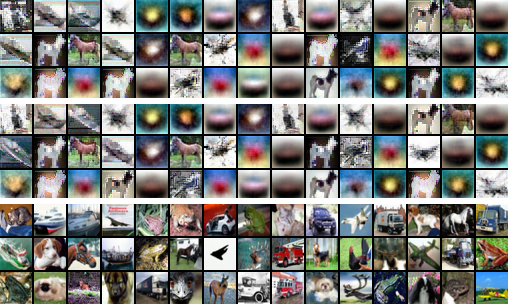}
\caption{Exact replica of Figure \ref{Fig.Results.1}, but for testing, instead of training observations.The \textbf{Bottom} shows the first 45 raw testing images from CIFAR-10. The \textbf{Middle} 45 images show \emph{reconstruction} of those raw images. No image has a good reconstruction! The \textbf{Top} 45 images show \emph{generation} using different than the resp. images in the \textbf{Middle} random sample but the same tentative empirical drift (from the second run of our construct).}
\label{Fig.Results.1a}
\end{figure*}

\bibliography{bibliography}
\bibliographystyle{icml2018}

\appendix
\section{Network architecture}

Network architecture is similar to the one used in \cite{Georgiev15-3} for CIFAR-10 in that it has class-dependent sampler and decoder. Optimizer is Adam,  \cite{Kingma15}, learning rate = 0.001, learning rate period of half-decay = 100 epochs, batch size = 1,000. Only one observation per minibatch is well reconstructed. We hard-coded training observations, number 8, 1020, 2016 in the initial CIFAR-10 order, to be well reconstructed (see Figures \ref{Fig.Results.0}, \ref{Fig.Results.2} and \ref{Fig.Results.3}). The other $47=50 - 3$ well reconstructed observations were chosen by the weight classifier.

Weight initialization for VAE is random normal, re-scaled for asymptotic behavior of its maximum eigenvalue, as in \cite{Georgiev15-1}, sub-section 6.1. Laplacian pyramid, \cite{Denton15}, is used once from 32 x 32 pixels to 16 x 16 pixels, using max-pooling when downsampling and repeats when upsampling back up. Weight sizes (with 10 classes) for VAE of the top Laplacian pyramid layer are 3*32*32-16-(2*10)-(16*10)-(3*32*32*10) and for the bottom Laplacian pyramid layer 3*16*16-4-(2*10)-(4*10)-(3*16*16*10). Only bottom Laplacian pyramid layer uses generative weights, 2 per class. The second term (when minimizing the negative log-likelihood) due to Laplacian pyramid, seem to act as a regularizer: much larger networks were tried, achieving the same results (we tried weight sizes for the top Laplacian pyramid layer 3*32*32-2048-(2*10)-(2048*10)-(3*32*32*10) and for the bottom Laplacian pyramid layer 3*16*16-1024-(2*10)-(1024*10)-(3*16*16*10)).

The more dimensions of generative weights are used in latent layer, the harder it becomes to replicate human creativity.

CNN with fully-connected layer at the end and initialized as uniform was used for weight classifier, with weight sizes 3-64-64-128-1. No dropout was used in classifier. The maximum ratio between weights was set to $10^6$ and the sum of all weights in a minibatch equals the size of the minibatch (the function \emph{softmax} was used at the end of the weight classifier).

The non-linearities used are $tanh$ in the decoder of VAE, its primitive -- $\ln cosh$ in the encoder of VAE (non-linearities are not used when computing drift or standard deviation) and usual ReLU in weight classifier.

We are random sampling from a non-Gaussian density, namely Laplacian\footnote{ Technically, Laplacian is not in the exponential/Gibbs class, but it is a sum of two exponential/Gibbs densities in the domains $(-\infty, \mu)$, $[\mu, \infty)$ defined by its mean $\mu$, and those densities are in the exponential/Gibbs class in their respective domains. Laplacian  is biologically-plausible because it is a  bi-product of squaring Gaussians.}. In order to have an unity variance, we choose for the independent one-dimensional latent  $p(z)$ $=p^{Lap}(z; 0,\sqrt{0.5})$, where $p^{Lap}(z; \mu,b)$ $=exp(-|z-\mu|/b)/(2b)$ is the standard Laplacian density with mean $\mu$ and scale $b$. In order to have zero generative error when $(\mu_{1,2}, \sigma_{1,2})$ $\rightarrow (0,1)$, we parametrize the conditional posterior as $p(z|.)$ $=p^{Lap}(z; \mu_{1,2},\sigma_{1,2} \sqrt{0.5})$. The generative error in (\ref{method.3}) equals:  $- \ln \frac{\sigma_2}{\sigma_1}$ $ + \frac{|\mu_1 - \mu_2|}{\sqrt{0.5}\sigma_2}$ $ + \frac{\sigma_1}{\sigma_2} exp\left(- |\mu_1 - \mu_2|/(\sigma_1\sqrt{0.5})\right ) - 1$,  see \cite{Gil13}, Table 3. In the first run $(\mu_2, \sigma_2)$ $= (0, 1)$ as usual, in the second run and onwards, the observation empical drift/standard deviation are used for $(\mu_2, \sigma_2)$.

Regular batch normalization, \cite{Ioffe15}, is used in encoder, decoder and weight classifier. In the presence of class-dependent decoder and batch normalization, there is a dependence on minibatch and class. Then, for hard-coded training images, sub-section \ref{training observation}, one has to use without any loss, for the whole respective training minibatch,  the class to which hard-coded images belong - see Figure \ref{Fig.Results.1}. For other types of dependence on minibatch, in the presence of batch normalization, see \cite{Ioffe17}.

\end{document}